# Identifying Mislabeled Training Data

**Carla E. Brodley**                                         BRODLEY@ECN.PURDUE.EDU
*School of Electrical and Computer Engineering*
*Purdue University*
*West Lafayette, IN 49707 USA*

**Mark A. Friedl**                                            FRIEDL@CRSA.BU.EDU
*Department of Geography and Center for Remote Sensing*
*675 Commonwealth Avenue*
*Boston University*
*Boston, MA 02215 USA*

## Abstract

This paper presents a new approach to identifying and eliminating mislabeled training instances for supervised learning. The goal of this approach is to improve classification accuracies produced by learning algorithms by improving the quality of the training data. Our approach uses a set of learning algorithms to create classifiers that serve as noise filters for the training data. We evaluate single algorithm, majority vote and consensus filters on five datasets that are prone to labeling errors. Our experiments illustrate that filtering significantly improves classification accuracy for noise levels up to 30%. An analytical and empirical evaluation of the precision of our approach shows that consensus filters are conservative at throwing away good data at the expense of retaining bad data and that majority filters are better at detecting bad data at the expense of throwing away good data. This suggests that for situations in which there is a paucity of data, consensus filters are preferable, whereas majority vote filters are preferable for situations with an abundance of data.

## 1. Introduction

One goal of an inductive learning algorithm is to form a generalization from a set of labeled training instances such that classification accuracy for previously unobserved instances is maximized. The maximum accuracy achievable depends on the quality of the data and on the appropriateness of the chosen learning algorithm for the data. The work described here focuses on improving the quality of training data by identifying and eliminating mislabeled instances prior to applying the chosen learning algorithm, thereby increasing classification accuracy.

Labeling error can occur for several reasons including subjectivity, data-entry error, or inadequacy of the information used to label each object. Subjectivity may arise when observations need to be ranked in some way such as disease severity or when the information used to label an object is different from the information to which the learning algorithm will have access. For example, when labeling pixels in image data, the analyst typically uses visual input rather than the numeric values of the feature vector corresponding to the observation. Domains in which experts disagree are natural places for subjective labeling errors (Smyth, 1996). In other domains, the most frequent type of error is mistakes made during data-entry. A third cause of labeling error arises when the information used to label each observation is inadequate. For example, in the medical domain it may not be possible to perform the tests necessary to guarantee that a diagnosis is 100% accurate. For domains in which labeling errors occur, an automated method of eliminating or correcting mislabeled observations will improve the predictive accuracy of the classifier formed from the training data.

In this article we address the problem of identifying training instances that are mislabeled. Quinlan (1986) demonstrated that as noise level increases, removing noise from attribute information decreases the predictive accuracy of the resulting classifier if the same attribute noise is present in the data to be classified. In the case of mislabeled training instances (class noise) the opposite is true; *cleaning* the training data will result in a classifier with significantly higher predictive accuracy. For example, Brodley and Friedl (1996a, 1996b) illustrated that for class noise levels of less than 40%, removing mislabeled instances from the training data resulted in higher predictive accuracy relative to classification accuracies achieved without "cleaning" the training data.

We introduce a method for identifying mislabeled instances that is not specific to any particular learning algorithm, but rather serves as a general method that can be applied to a dataset before feeding it to a learning algorithm. The basic idea is to use a set of learning algorithms to create classifiers that act as a *filter* for the training data. The method is motivated by the technique of removing *outliers* in regression analysis (Weisberg, 1985). An outlier is a case (an instance) that does not follow the same model as the rest of the data and appears as though is comes from a different probability distribution. Candidates are cases with a large residual error.[1] Weisberg (1985) suggests building a model using all of the data except for the suspected outlier and testing whether it does or does not belong to the model using the externally studentized $t$-test.

Here, we apply this idea by using a set of classifiers formed from part of the training data to test whether instances in the remaining part of the training data are mislabeled. An important difference between our work and previous approaches to outlier detection is that our approach assumes that the errors in the class labels are independent of the particular model being fit to the data. In essence, our method attempts to identify data points that would be outliers in *any* model.

We evaluate our approach on five datasets that are prone to labeling errors and we find that filtering substantially improves performance when labels are noisy. In addition, we compare filtering to majority vote ensemble classifiers to illustrate that although majority vote classifiers provide some protection against noisy data, filtering results in significantly higher accuracy. A third experiment evaluates the precision of our method in identifying only mislabeled data points. We conclude with a discussion of future research directions aimed at minimizing the probability of discarding instances that are exceptions rather than noise.

---

1. Not all residual cases are outliers because according to the model, large deviations will occur with the frequency prescribed by the generating probability distribution.

## 2. Related Work

The problem of handling noise has been the focus of much attention in machine learning and most inductive learning algorithms have a mechanism for handling noise in the training data labels. For example, pruning in decision trees is designed to reduce the chance that the tree is overfitting to noise in the training data. As pointed out by Gamberger, Lavrač and Džeroski (1996), removing noise from the data before hypothesis formation has the advantage that noisy examples do not influence hypothesis construction.

The idea of eliminating instances to improve the performance of nearest neighbor classifiers has been a focus of research in both pattern recognition and instance-based learning. Wilson (1972) used a $k$-nearest neighbor ($k$-NN) classifier (in experiments $k$ was set to three) to select instances that were then used to form a 1-NN classifier; only instances that the $k$-NN classified correctly were retained for the 1-NN. Tomek (1976) extended this approach with a procedure that calls Wilson's algorithm for increasing values of $k$. Wilson and Martinez (1997, 1999) have incorporated this approach into a suite of instance selection techniques for exemplar-based learning algorithms. Aha, Kibler and Albert (1991) demonstrated that selecting instances based on records of their contribution to classification accuracy in an instance-based classifier improves the accuracy of the the the resulting classifier. Skalak (1994) created an instance selection mechanism for nearest neighbor classifiers with the goal of reducing their computational cost, which depends on the number of stored instances. The selection of a few instances (designated as prototypes) by a Monte Carlo sampling algorithm demonstrated that accuracy was maintained and even raised for several data sets. Wilson (1999) provides a comprehensive overview of instance selection techniques for exemplar-based learning algorithms.

The idea of selecting "good" instances has also been applied to other types of classifiers. Winston (1975) demonstrated the utility of selecting "near misses" when learning structural descriptions. Skalak and Rissland (1990) describe an approach to selecting instances for a decision tree algorithm using a case-based retrieval algorithm's taxonomy of cases (for example "the most-on-point cases"). Lewis and Catlett (1994) illustrate that sampling instances using an estimate of classification certainty drastically reduces the amount of data needed to learn a concept.

A danger in automatically removing instances that cannot be correctly classified is that they might be exceptions to the general rule. When an instance is an exception to the general case, it can appear as though it is incorrectly labeled. A key question in improving data quality is how to distinguish exceptions from noise. Guyon, Matic and Vapnik's (1996) approach uses an information criterion to measure an instance's typicality; atypical instances are then presented to a human expert to determine whether they are mislabeled or exceptions. However, they note that because their method is an on-line method it suffers from ordering effects. Oka and Yoshida (1993, 1996) created a method that learns generalizations and exceptions separately by maintaining a record of the correctly and incorrectly classified inputs in the influence region of each stored example. The mechanism for distinguishing noise from exceptions is based on a user-specified parameter, which is used to ensure that each stored sample's classification rate is sufficiently high. To our knowledge, the approach has only been tested on artificial datasets.

Srinivasan, Muggleton and Bain (1992) use an information theoretic approach to detect exceptions from noise during the construction of a logical theory. Their motivation is that there is no mechanism by which a non-monotonic learning strategy can reliably distinguish true exceptions from noise. Methods based on closed word specialization (Bain & Muggleton, 1991) overfit the data. To select the next clause to add to the current theory, they select the one that corrects the most errors (they found empirically that the more robust method that also considers the complexity of the clause does not impact results). To address the problem that the best clause may not produce an immediate increase in compression, they continue to add clauses, waiting to make the update final until they obtain a compression. This can occur after several clauses have been added. If compression never comes, then the clause (and subsequent clauses) are not added to the theory. Their method is analogous to pre-pruning of decision trees. In their experiments, they injected random classification noise (Angluin & Laird, 1988) into the data. This is identical to our method for injecting noise for *two class cases*. For multiclass cases, our experimental method injects noise in the manner that it would naturally occur in the domain (see Section 4).

Gamberger and Lavrač (1996) and Gamberger, Lavrač and Džeroski (1996) have developed a method for handling noise that first removes inconsistent examples from the training data. Inconsistent examples are those that have the same values for the features but different class labels. They then transform the features into a binary feature set. Next they examine which set of examples, when removed, reduces the total number literals needed to retain the property that the current set of instances is not inconsistent. They have a user-set threshold that monitors how big this example set should be. Given two sets of examples that result in an equal reduction in the amount of literals, we would like to select the smaller based on the heuristic that it is more likely to be noise.

Zhao and Nishida (1995) deal with a related issue – the problem of noise in feature measurements. Their approach extends fuzzy logic's approach to representation and calculation of inaccurate data. They identify inaccurate data on the basis of qualitative correlations among related data based on the observation that some features are qualitatively dependent such as symptomatic data reflecting a patients disease. For example if $n-1$ our of $n$ symptoms indicate that a patient has a particular disease, then we might believe that the value of $n^{th}$ symptom was incorrectly measured or entered. Their method dynamically determines fuzzy intervals for inaccurate data and requires that they have domain knowledge to divide the features into sets whose members are qualitatively dependent. When no domain knowledge is available, they suggest using a fuzzy logic system that has predetermined intervals for the features.

Several recent developments have greatly helped with learning exceptions even in the face of noisy data. Dietterich and Bakiri (1995) developed a method for learning classifiers for multiple classes in which error-correcting output codes are employed as a distributed output representation (each class is assigned a unique binary string of length $n$). They illustrated that classification can be viewed as a communication problem in which the identity of the correct output class for a new example is being "transmitted" over a noisy channel. An empirical evaluation demonstrated that error-correcting codes can be used to improve performance. Another recent innovation is boosting (Schapire, 1990; Quinlan, 1996), which forms a set of classifiers whose predictions are combined by voting. Boosting adjusts the weights of the training samples at each iteration, paying more attention to samples that are "difficult to learn." One potential problem with these methods is that they may generate classifiers that have been fit to systematic noise. Moreover, this situation may be difficult to detect. Typically, to validate a particular classifier or method, one does a cross-validation over a set of labeled data. If all of the data was labeled by the same mechanism then the entire dataset contains the same systematic errors. Achieving high accuracy on this type of "independent" test set, may mean that the method has done an excellent job at fitting to the systematic noise.

In the past decade, the computational machine learning community has investigated variations of PAC learning that model the type of noise that might occur in a real learning environment (Angluin & Laird, 1988; Sloan, 1988; Decatur, 1996). More recently, models of non-uniform classification noise (Sloan, 1995) and partial non-uniform noise (Decatur, 1997) have been introduced. These models do not assume that each instance has the same misclassification rate and therefore are more realistic models of the types of noise observed in real-world applications. A recent innovation is to alter the learning procedure for the known noise rates (Decatur, 1997). However in most real-world scenarios one will not have access to the true noise rates of the various classes. In these cases, Decatur (1997) suggests searching for the noise rate using a cross-validation search, but this approach assumes that one has noise free data with which to evaluate the results of the search.

## 3. Filtering Training Data

This section describes a general procedure for identifying mislabeled instances in a training set. The first step is to identify candidate instances by using $m$ learning algorithms (called *filter algorithms*) to tag instances as correctly or incorrectly labeled. To this end, a $n$-fold cross-validation is performed over the training data. For each of the $n$ parts, the $m$ algorithms are trained on the other $n-1$ parts. The $m$ resulting classifiers are then used to tag each instance in the excluded part as either correct or mislabeled. An individual classifier tags an instance as mislabeled if it classifies the instance as belonging to a different class than that given by its training label. Note that when $n$ is equal to the total number of training instances, this method differs from Weisberg's (1985) outlier detection method only in the test used to determine whether a case is an outlier.

At the end of the $n$-fold cross-validation each instance in the training data has been tagged. Using this information, the second step is to form a classifier using a new version of the training data for which all of the instances identified as mislabeled are removed. Filtering can be based on one or more of the $m$ base level classifiers' tags. The filtered set of training instances is provided as input to the *final learning algorithm*. The resulting classifier is the end product of the approach. Figure 1 depicts the general procedure. Specific implementations of this general procedure differ in how the filtering is performed, and in the relationship between the filter algorithm(s) and the final learning algorithm(s).

### 3.1 Single Algorithm Filters

One approach is to use the same learning algorithm to construct both the filter and the final classifier. This approach is most similar to removing outliers in regression analysis, for which the same model is used to test for outliers and for fitting the final model to the data once the outliers have been removed. A related method is that proposed by John (1995) for removing the training instances that are pruned by C4.5 (Quinlan, 1993). Specifically, for each leaf node in the pruned tree for which training instances are observed from more than one class, John's method eliminates those instances that are not from the majority class. The tree is then rebuilt from the reduced set of training instances. This process iterates until no further pruning can be done. The key difference between our method and John's is that our method uses a cross-validation over the training data with one iteration whereas John's method deals with the training examples directly and performs multiple iterations.

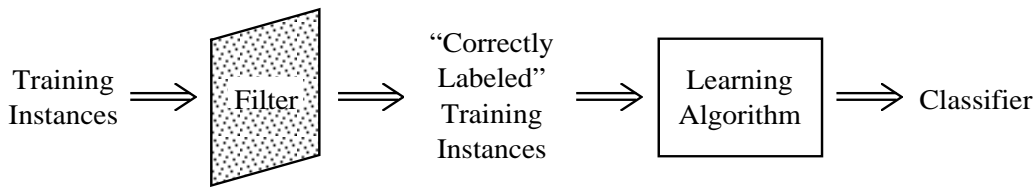

Figure 1: The general procedure for eliminating mislabeled instances.

A second way to implement filtering is to construct a filter using one algorithm and to construct the final classifier using a different algorithm. The assumption underlying this approach is that some algorithms act as good filters for other algorithms, much like some algorithms act as good feature selection methods for others (Cardie, 1993). The approach described by Wilson (1972) to filtering data for a 1-NN using a $k$-NN is an example of this approach.

## 3.2 Ensemble Filters

Ensemble classifiers combine the outputs of a set of base-level classifiers (Hansen & Salamon, 1990; Benediktsson & Swain, 1992; Wolpert, 1992). A majority vote ensemble classifier will outperform each base-level classifier on a dataset if two conditions hold: (1) the probability of a correct classification by each individual classifier is greater than 0.5 and (2) the errors in predictions of the base-level classifiers are independent (Hansen & Salamon, 1990).

In filtering, an ensemble classifier detects mislabeled instances by constructing a set of *base-level detectors* (classifiers) and then using their classification errors to identify mislabeled instances. The general approach is to tag an instance as mislabeled if $x$ of the $m$ base-level classifiers cannot classify it correctly. In this work we examine both majority and consensus filters. A majority vote filter tags an instance as mislabeled if more than half of the $m$ base level classifiers classify it incorrectly. A consensus filter requires that *all* base-level detectors must fail to classify an instance as the class given by its training label for it to be eliminated from the training data.

It is important to note that the underlying premise of an ensemble filter differs from methods developed in regression analysis, in which outliers are defined relative to a particular model. Here we assume that some instances in the data have been mislabeled and that the label errors are independent of the particular model being fit to the data. Therefore collecting information from different models will provide a better method for detecting mislabeled instances than collecting information from a single model.

Scarcity of training data is a problem in many classification and learning problem domains (e.g., medical diagnosis). For such datasets, we want to minimize the probability of discarding an instance that is an exception rather than an error. Indeed, Danyluk and Provost (1993) note that learning from noisy data is difficult because it is hard to distinguish between noise and exceptions, especially if the noise is systematic. Ideally, the biases of at least one of the learning algorithms will enable it to learn the exception. Therefore, one or more of the classifiers that comprise the base-level set of detectors can have difficulty capturing a particular exception without causing the exception to be erroneously eliminated from the training data. In this case, the consensus method will make fewer detection errors than a majority or single algorithm method. Taking a consensus rather than a majority vote is a more conservative approach and will result in fewer instances being eliminated from the training data. The drawback of a conservative approach is the added risk of retaining bad data. In the next section we analyze the probabilities of making identification errors for both retaining bad data and throwing away good data for majority and consensus filters.

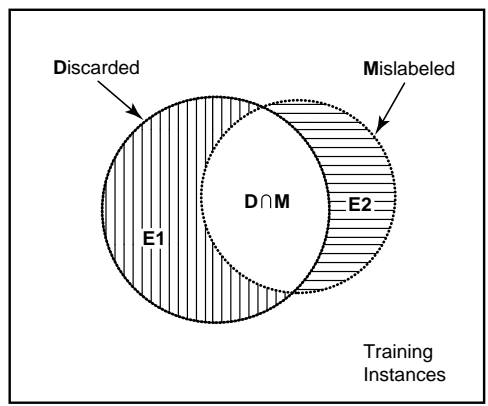

Figure 2: Types of detection errors

## 3.3 Identification Errors

In identifying mislabeled instances there are two types of error that can be made (see Figure 2). The first type ($E1$) occurs when an instance is incorrectly tagged as mislabeled and is subsequently discarded (D). The second type of error ($E2$) occurs when a mislabeled instance (M) is tagged as correctly labeled. In this section we analyze the probability of each of these types of errors for the consensus and majority filter methods.

### 3.3.1 MAJORITY VOTE

The event of incorrectly tagging a correct instance as mislabeled happens when more than half of the $m$ base-level detectors fail to classify the instance correctly. Let $P(E1_i)$ be the probability that classifier $i$ makes an $E1$ error and for the sake of clarity assume that all $m$ base-level classifiers have the same probability of making an $E1$ error that is equal to $P(E1_i)$. If we assume that the errors of the base-level classifiers are independent, the probability that a majority vote filter will throw out good data is given by:

$$P(E1) = \sum_{j>m/2}^{j=m} P(E1_i)^j (1 - P(E1_i))^{m-j} \left( \begin{array}{c} m \\ j \end{array} \right)$$

Where $P(E1_i)^j (1 - P(E1_i))^{m-j} \begin{pmatrix} m \\ j \end{pmatrix}$ represents the chance of $j$ errors among the $m$ base-level classifiers. If the probability of making an $E1$ error is less than 0.5, then the majority filter will make fewer errors than a single-algorithm filter formed from one of the base-level classifiers.

The probability of mistaking a mislabeled instance for a correctly labeled instance ($E2$) occurs when more than half of the base-level classifiers classify the instance as the mislabeled class.[2] Let $P(E2_i)$ be the probability that a base-level detector $i$ makes an error of type $E2$. Assuming that the errors are independent and that the probabilities of the base-level classifiers making an $E2$ error are the same, then the probability that the majority vote filter makes a type $E2$ error is given by

$$P(E2) = \sum_{j > m/2}^{j=m} P(E2_i)^j (1 - P(E2_i))^{m-j} \begin{pmatrix} m \\ j \end{pmatrix}$$

Therefore, a majority vote filter will make fewer $E2$ errors than a single-algorithm filter if $P(E2_i)$ is less than 0.5. When all base-level classifiers' $E2$ errors are made on the same subset of the instances, the probability that a majority vote classifier will make an $E2$ error is identical to the probability that a single-algorithm filter will make an error.

### 3.3.2 Consensus Filters

For a consensus filter, an $E1$ error occurs when all of the base-level detectors fail to classify an instance correctly. Let $P(E1_i)$ be the probability that base-level detector $i$ makes an $E1$ error and $m$ be the number of base-level classifiers, then the general form of the probability of making an $E1$ error is given by:

$$P(E1) = P(E1_1)P(E1_2 \mid E1_1)...P(E1_n \mid E1_1 \cap \cdots E1_{m-1})$$

If the base-level detectors make errors on the same instances then the probability of an $E1$ error is equal to the probability that a single base-level detector makes an error $P(E1_i)$. When the $E1$ errors of the base-level detectors are independent, then a consensus filter has a smaller probability of making an $E1$ error than each of its base-level detectors and the probability of making an $E1$ error is given by:

$$P(E1) = \prod_{i=1}^{m} P(E1_i)$$

If the assumption of independence of the errors of the base-level detectors holds, then we would expect a consensus filter to have a smaller probability of making an $E1$ error than a single-algorithm filter.

---

2. For the multiclass case, a majority can be fewer than one half of the base level classifiers. Because our analysis defines an $E2$ error to be when more than half of the base level classifiers make an $E2$ error, it is an underestimate for the multiclass case. Note that since our empirical analysis is based on the results of ensembles containing three base level classifiers, the probability of an $E2$ error for the two class case and the multiclass case are computed the same way.

The probability of mistaking a mislabeled instance for a correctly labeled instance ($E2$) occurs when an instance is mislabeled and one or more of the base-level classifiers predicts the mislabeled class. Let $P(E2_i)$ be the probability that a base-level detector $i$ makes an error of type $E2$. A consensus filter makes a type $E2$ error if one or more of the base-level classifiers makes a type $E2$ error. This probability is equal to one minus the probability that none of the base-level detectors makes an $E2$ error. If $1 - P(E2_i)$ is the probability that classifier $i$ does not make an $E2$ error, then the probability that a consensus filter makes an $E2$ error is given by:

$$P(E2) = 1 - (1 - P(E2_1))(1 - P(E2_2 \mid \overline{E2_1}))...(1 - P(E2_m \mid \overline{E2_1} \cap ...\overline{E2_{m-1}}))$$

When the probability of a base-level classifier making an $E2$ error is independent of the probability of the other base-level classifiers making an $E2$ error this becomes:

$$P(E2) = 1 - \prod_{i=1}^{m}(1 - P(E2_i))$$

Therefore, in direct contrast to $E1$ errors, independence of the $E2_i$ errors can lead to higher overall $E2$ error for the consensus filter. In such cases, a single-algorithm filter would make fewer $E2$ errors than a consensus filter that contains the single algorithm as one of its base-level classifiers.

### 3.4 Mislabeled Instances versus Exceptions

Before moving on to an evaluation of the approach, the issue must be addressed that instances tagged as mislabeled by the above approach could be exceptions to a general rule and therefore would need special treatment. When an instance is an exception to the general case, it can appear as though it is incorrectly labeled. When applying techniques to identify and eliminate noisy instances, one wants to avoid discarding correctly labeled exceptions. In Section 5 we discuss future plans for learning to distinguish noise from exceptions.

A second situation in which an instance might be discarded erroneously by our filter approach is if an algorithm with an inappropriate learning bias for the data set is used. In such cases, the algorithm's representation language may not permit an accurate representation of the concept. This problem is analogous to situations in which removing outliers does little to improve the fit of a first-order linear regression if the correct model of the data is quadratic. Finally, since the filter algorithm(s) constructs a classifier using the original noisy data set, the identification of mislabeled instances is bound to include errors; using a classifier formed from mislabeled instances to determine if other instances are mislabeled will lead to some errors. With these caveats in mind, we now proceed to an empirical evaluation of the approach.

## 4. Empirical Evaluation

To evaluate the ability of the various filtering approaches to identify mislabeled training instances we chose domains for which labeling error occurs naturally. To simulate the types of error that occur in practice, we consulted domain experts for each dataset to identify the pairs of classes likely to be confused. To test the filtering approach we artificially introduced noise into the training labels between these pairs of classes. We did not introduce noise between all pairs of classes as this would not model the types of labeling errors that occur in practice. Our experiments are designed to assess the different types of filters' ability to identify mislabeled instances and the effect that eliminating mislabeled instances has on predictive accuracy. We describe our experimental method in Section 4.2.

| | Class Name | Instances |
|---|---|---|
| 1 | broadleaf evergreen forest | 628 |
| 2 | coniferous evergreen forest & woodland | 320 |
| 3 | high latitude deciduous forest & woodland | 112 |
| 4 | tundra | 735 |
| 5 | deciduous-evergreen forest & woodland | 57 |
| 6 | wooded grassland | 212 |
| 7 | grassland | 348 |
| 8 | bare ground | 291 |
| 9 | cultivated | 527 |
| 10 | broadleaf deciduous forest & woodland | 15 |
| 11 | shrubs and bare ground | 153 |

Table 1: Land cover classes

## 4.1 Domains

This research originated from efforts addressing the task of automated land-cover mapping from satellite data. In applying machine learning techniques to this problem we developed the idea of using consensus filters to remove mislabeled training instances. Results from this work can be found in (Brodley & Friedl, 1996a, 1996b). To explore this question further, we chose four additional datasets – our choice was based on a judgment of whether the labeling process included substantial levels of subjectivity or noise. In this section, we identify how labeling errors may arise in each of the five domains.

### 4.1.1 Automated Land Cover Mapping

The first dataset we examined consists of a time series of globally distributed satellite observations of the Earth's surface. The dataset was compiled by DeFries and Townshend (1994), and includes 3398 locations that encompass all major terrestrial biomes[3] and land cover types (see Table 1).

The remote sensing observations are measurements of a parameter called the normalized difference vegetation index (NDVI). This index is commonly used to infer the amount of live vegetation present within a pixel at the time of data acquisition. The NDVI data used here were collected by the Advanced Very High Resolution Radiometer on board the National Oceanic and Atmospheric Administration series of meteorological satellites. The data have been re-sampled from the raw satellite data to have a uniform spatial resolution of one degree of latitude and longitude. Each one degree pixel is described by a time series of twelve NDVI values at monthly time increments from 1987, and by its latitude, which can be useful for discriminating among classes with otherwise similar spectral properties (Defries & Townshend, 1994). The monthly temporal sampling procedure allows the compilation of cloud-free views of the Earth's surface, and also captures seasonal dynamics in vegetation. The temporal information is particularly useful for classification of vegetation and land cover, as seasonal changes in vegetation are one of the best indicators of vegetation type.

---

3. A *biome* is the largest subdivision of the terrestrial ecosystems. Some examples of biomes are grasslands, forests and deserts.

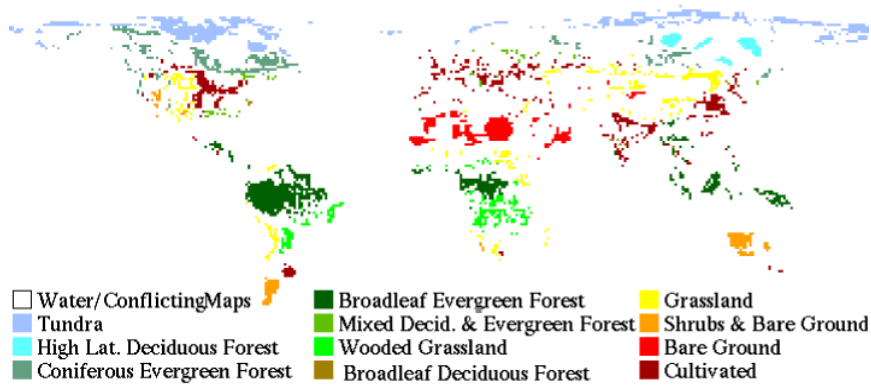

Figure 3: Training and testing sites.

A summary of these data is provided in Table 1. The class labels were selected to reflect fairly broad classes with extensive geographic coverage. Maps developed from this classification scheme may then be used to relate land cover classes to structurally and functionally significant ecological properties. From a remote sensing perspective, this classification system reflects a compromise between class labels that are separable from coarse resolution remote sensing data, and class labels that are useful to end-users such as ecologists. For further details regarding the specific procedures used to compile the data, the reader is referred to DeFries and Townshend (1994) and Los, Justice and Tucker (1994).

Labeling error occurs in land-cover training data for many reasons. One source arises because discrete classes and boundaries are used to distinguish among classes that have subtle boundaries in space and that have fairly small differences in terms of their physical attributes. For example, the distinction between a grassland and wooded grassland can be quite difficult to discern. Consequently, pixels labeled as grassland may in fact represent open woodland areas and vice versa. This source of error is especially problematic at the one degree spatial resolution of the data used here. Another source of error is the difference between potential and actual vegetation. Potential vegetation refers to the type of vegetation that occurs naturally in a region based on soil, climate and geologic controls. Actual vegetation refers to the vegetation present in the region. Differences arise because humans have substantially modified the Earth's surface from its natural state. Labeling error often occurs because potential vegetation labels are used in the absence of other information.

Another source of error arises because of land-cover change. In areas undergoing rapid economic development (e.g., the humid subtropics) information quickly becomes out of date. These problems are best illustrated by the source of our data, which come from three existing maps of global land cover (Matthews, 1983; Olson, Watts, & Allison, 1983; Wilson & Henderson-Sellers, 1985). Comparison of land cover labels among the three maps shows that they agree for only approximately 20% of the Earth's land surfaces. (See Figure 3 for locations where the three maps were in agreement.) The problem of using data collected from multiple experts has been documented in other domains as well (Smyth, 1996).

For this work, based on our expert's suggestions, we introduced random error between the following pairs of classes: 3-4, 5-2, 6-7, 8-11, 5-10 (see Table 1 for the names of the classes) (Brodley & Friedl, 1996a).

### 4.1.2 CREDIT APPROVAL

The goal of credit approval is to determine whether to give an applicant a credit card. Our dataset includes 690 instances labeled as positive or negative. There are nine discrete attributes with two to fourteen values, and six continuous attributes. One or more attribute values are missing from 37 instances. The class distribution is fairly well balanced, with 307 instances labeled "+" and 383 instances labeled "-".

This domain was chosen because the choice of whether to give an applicant credit is subjective in nature. Error is introduced because an assessment of future behavior is based on past behavior. In an identical application (but using a different dataset), American Express-UK found that loan officers were less than 50% correct at predicting whether "borderline" applicants would default on their loans (Michie, 1989). This means that 50% of the labels were in error or the attributes were not adequate to distinguish good from bad applicants. (It is interesting to note that a decision tree was able to classify 70% of borderline applicants correctly in the UK American Express domain (Michie, 1989)).

### 4.1.3 SCENE SEGMENTATION

For this data set, the goal is to learn to segment an image into the seven classes: sky, cement, window, brick, grass, foliage and path. Each of the classes has 330 observations, yielding 2310 total observations. Each instance is the average of a $3 \times 3$ window of pixels represented by 17 low-level, real-valued image features. The instances were drawn randomly from a database of seven outdoor images from buildings around the University of Massachusetts at Amherst campus.

The labels for this dataset were produced by first running the images through a color segmentation algorithm (the NKGB algorithm (Draper, Collins, Brolio, Hanson, & Riseman, 1989)) and then manually labeling each region on a computer monitor. This procedure produces two types of labeling errors: objects that blend into one another, such that a region that is predominantly one type of object has pixels from another object class in it; and regions for which the boundary is unclear, even for visual inspection by humans (Draper, 1998). For example, because sky tends to "poke through" foliage, sky and foliage can be confused in the training data. In the experiments that follow we introduced the following confusions: sky-foliage; path-grass; grass-foliage.[4]

---

4. These confusions were suggested by Bruce Draper, who is the creator of the database.

|   | Class Name | Insts |
|---|------------|-------|
| 1 | Road       | 238   |
| 2 | Roadline   | 21    |
| 3 | Gravel     | 42    |
| 4 | Grass      | 63    |
| 5 | Dirt       | 9     |
| 6 | Foliage    | 1444  |
| 7 | Trunk      | 185   |
| 8 | Sky/Tree   | 22    |
| 9 | Sky        | 32    |

Table 2: Road segmentation classes

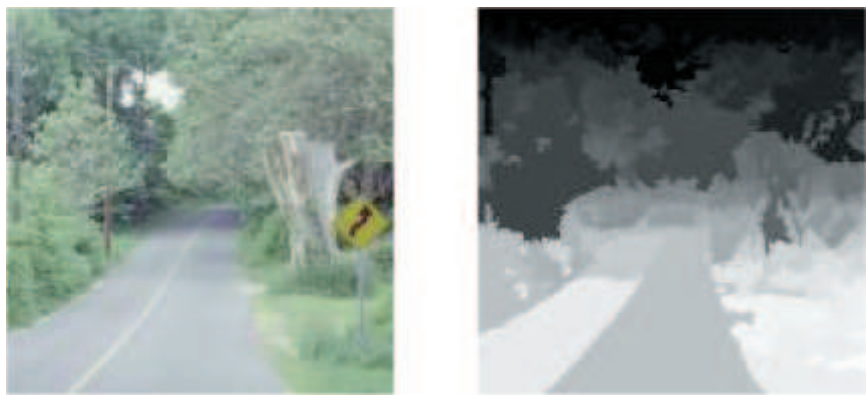

Figure 4: Road dataset: original and segmented training image.

### 4.1.4 ROAD SEGMENTATION

This data comes from a set of images of country roads in Massachusetts. Each instance represents a $3 \times 3$ grid of pixels described by three color and four texture features. The classes are road, roadline, gravel, grass, dirt, foliage, trunk, sky/tree and sky. There are 2056 instances in this data set and 105 attribute values are missing. The frequency distribution of classes is shown in Table 2. The labeling procedure for this domain was the same as for the scene segmentation domain. Figure 4 shows an original image on the left and its corresponding segmentation by the NKBG algorithm on the right, which is used by a human to create training labels. It is clear from this example, that the regions produced by the segmentation algorithm are noisy in nature – for example locating all of the tree trunk on the right of the image using the regions produced by the NKGB algorithm is impossible.

In our experiments we introduced the following confusions: foliage-sky (6-9); gravel-dirt (3-5); grass-dirt (4-5); road-gravel (1-3); grass-foliage (4-6); sky-sky/tree (9-8); foliage-sky/tree(6-8) and foliage-trunk(6-7).[5]

---

5. These confusions were suggested by Bruce Draper, who is the creator of the database.

### 4.1.5 FIRE DANGER PREDICTION

The goal of this dataset is to rank fires in terms of their severity on a scale of 1-8. A label of 1 indicates no fire occurred on that day and labels 2-8, represent fires from minor (2) to severe (8). The ranking of any event by humans is necessarily subjective in nature. Furthermore, since fires are events that occur over time, the chance of inconsistent ranking increases with the chance that a different person may do the ranking.

This dataset was compiled on bush fire activity in the mallee vegetation area of northwest Victoria, Australia (Dowe & Krusel, 1993a, 1993b, 1994). The dataset contains 3833 observations (days) each described by ten features that measure the maximum temperature, the number of days since it last rained, a drought index, the temperature at 3pm, the wet-bulb temperature, the wind speed, the relative humidity, the forest fire danger index, the air pressure, and the grass fire danger. In our experiments we tried to predict the exact fire severity, whereas historically the most common use of this data is to turn it into a binary prediction problem that distinguishes between no fire (class 1) and fire (classes 2-8). Because it seems highly unlikely that one would erroneously label a fire free day as having a fire and vice versa, we introduced class confusions among the pairs: 2-3, 3-4, 4-5, 5-6, 6-7, and 7-8.

## 4.2 Experimental Method

As described above, to test the single-algorithm, majority vote and consensus procedures, we introduced random noise into the training data between pairs of classes that are most likely to be confused in the original labels. In this way, we have simulated the type of labeling error that is common to each domain. The pairs of classes to which error was introduced for each domain were described in the previous section.

For each of ten runs, each dataset was randomly divided into a training (90%) set and a testing (10%) set. After the data was split into independent train and test sets, we then corrupted the training data by introducing labeling errors using noise levels ranging from 0 to 40% noise. For a noise level $x$, an individual observation whose class was one of the identified problematic pairs had an $x\%$ chance of being corrupted. For example, in the landcover domain an instance from class 8 (bare ground) has an $x\%$ chance of being changed to class 11 (shrubs and bare ground), and an instance from class 11 has an $x\%$ chance of being changed to class 8. Using this method the percentage of the entire training set that was corrupted may be less than $x\%$ for multi-class problems because only some pairs of classes are considered problematic. The actual percentage of noise in the corrupted training data is reported in tables that present the experimental results.

For each noise level, we compared the average predictive accuracy of classifiers trained using filtered versus unfiltered data. For each of the ten runs that make up the average, we used a four-fold cross-validation to filter the corrupted instances from the training data. To assess the ability of the single-algorithm, majority and consensus filter methods to identify the corrupted instances we then ran each of the learning algorithms twice: first using the unfiltered dataset then using the filtered dataset.

### 4.3 Learning Algorithms

We chose three well-known algorithms from the machine learning and statistical pattern recognition communities to form the filters: decision trees, nearest neighbor classifiers and linear machines. We restrict the presentation of our empirical results to these three algorithms to enhance the clarity of our presentation of the method and to reduce the number of tables presented. However, in addition to the experiments reported, we also ran experiments with five base level algorithms (the three in this paper plus 5-NN and LMDT (Brodley & Utgoff, 1995)). Our results for the increased set of learning algorithms showed the same trends as those reported in this article.

**A univariate decision tree (D-Tree)** is either a leaf node containing a classification or an attribute test, with for each value of the attribute, a branch to a decision tree. To classify an instance using a decision tree, one starts at the root node and finds the branch corresponding to the value of the test attribute observed in the instance. This process repeats at the subtree rooted at that branch until a leaf node is reached. The instance is then assigned the class label of the leaf. One well-known approach to constructing a decision tree is to grow a tree until each of the terminal nodes (leaves) contain instances from a single class and then prune back the tree with the objective of finding the subtree with the lowest misclassification rate. Our implemented algorithm uses C4.5's pruning method with a confidence level of 0.10 (Quinlan, 1993).[6]

To select a test for a node in the tree, we choose the test that maximizes the information-gain ratio metric (Quinlan, 1986). Our implementation sets the minimum number of instances to form a test node to be equal to two. Univariate decision tree algorithms require that each test have a discrete number of outcomes. To meet this requirement, each ordered feature $A_i$ is mapped to a set of unordered features by finding a set of Boolean tests of the form $A_i > b$, where $b$ is in the observed range and is a cut point of $A_i$. Our algorithm finds the value of $b$ that maximizes the information-gain ratio. To this end, the observed values for $A_i$ are sorted, and the midpoints between class boundaries are evaluated (Quinlan, 1986; Fayyad & Irani, 1992).

**A k-nearest neighbor (k-NN) classifier** (Duda & Hart, 1973) is a set of $n$ instances, each from one of m classes, that are used to classify an unlabeled instance according to the majority classification of the instance's $k$ nearest neighbors. In this version of the algorithm each instance in the training data presented to the algorithm is retained. To determine the distance between a pair of instances we apply the Euclidean distance metric. In our experiments $k$ was set to one.

**A linear machine (LM)** is a set of $R$ linear discriminant functions that are used collectively to assign an instance to one of the $R$ classes (Nilsson, 1965). Let $\mathbf{Y}$ be an instance description (a pattern vector) consisting of a constant 1 and the $n$ features that describe the instance. Then each discriminant function $g_i(\mathbf{Y})$ has the form $\mathbf{W}_i^T\mathbf{Y}$, where $\mathbf{W}_i$ is a vector of $n + 1$ coefficients. A linear machine infers instance $\mathbf{Y}$ belongs to class $i$ if and only if ($\forall j, j \neq i$) $g_i(\mathbf{Y}) > g_j(\mathbf{Y})$. For the rare cases in which $g_i(\mathbf{Y}) = g_j(\mathbf{Y})$ an arbitrary decision is made: our implementation of an LM chooses the smaller of $i$ and $j$ in these cases.

---

6. We chose a value of 0.10 because preliminary experiments indicated that this value performed best across all domains. Future work will examine the relationship between the pruning confidence level and the level of noise in the data.

| Noise Level | | 0 | 5 | 10 | 20 | 30 | 40 |
|---|---|---|---|---|---|---|---|
| Actual Noise | | 0.0 | 3.5 | 6.8 | 14.2 | 21.7 | 29.4 |
| 1-NN | None | $87.3 \pm 1.7$ | $84.2 \pm 1.7$ | $81.8 \pm 2.5$ | $76.5 \pm 1.8$ | $71.9 \pm 2.7$ | $67.3 \pm 3.0$ |
| | SF | $87.0 \pm 1.7$ | $87.2 \pm 2.0$ | $86.1 \pm 2.1$ | $82.4 \pm 1.8$ | $77.6 \pm 2.7$ | $72.5 \pm 2.8$ |
| | MF | $87.2 \pm 1.5$ | $87.3 \pm 1.6$ | $86.6 \pm 1.8$ | $85.0 \pm 1.7$ | $80.2 \pm 1.9$ | $75.0 \pm 3.9$ |
| | CF | $87.5 \pm 1.7$ | $87.4 \pm 1.6$ | $86.1 \pm 2.3$ | $82.8 \pm 1.9$ | $77.8 \pm 2.7$ | $72.9 \pm 3.0$ |
| LM | None | $78.6 \pm 2.2$ | $77.4 \pm 2.5$ | $77.5 \pm 2.4$ | $68.3 \pm 5.3$ | $68.9 \pm 9.2$ | $63.6 \pm 7.3$ |
| | SF | $79.1 \pm 1.5$ | $79.0 \pm 1.6$ | $78.2 \pm 2.5$ | $77.8 \pm 3.1$ | $74.0 \pm 3.5$ | $71.0 \pm 4.6$ |
| | MF | $79.5 \pm 1.9$ | $80.2 \pm 1.8$ | $79.7 \pm 2.4$ | $78.6 \pm 3.3$ | $76.4 \pm 3.9$ | $71.8 \pm 3.4$ |
| | CF | $80.0 \pm 2.1$ | $79.2 \pm 2.0$ | $79.6 \pm 2.0$ | $77.7 \pm 3.2$ | $74.3 \pm 4.3$ | $70.7 \pm 5.9$ |
| D-Tree | None | $85.6 \pm 1.7$ | $83.1 \pm 1.8$ | $80.8 \pm 2.0$ | $75.9 \pm 1.3$ | $69.9 \pm 1.8$ | $67.0 \pm 2.7$ |
| | SF | $84.7 \pm 1.7$ | $83.8 \pm 1.5$ | $84.5 \pm 1.6$ | $82.2 \pm 2.2$ | $79.6 \pm 2.2$ | $71.9 \pm 3.6$ |
| | MF | $84.7 \pm 1.6$ | $83.3 \pm 2.0$ | $83.9 \pm 1.7$ | $83.5 \pm 2.1$ | $79.4 \pm 2.5$ | $73.5 \pm 3.4$ |
| | CF | $85.5 \pm 2.6$ | $85.8 \pm 1.9$ | $85.2 \pm 1.6$ | $82.8 \pm 1.4$ | $78.1 \pm 2.5$ | $71.3 \pm 3.1$ |

Table 3: Classification accuracy – land cover data

To find the weights of the linear machine we use the thermal training rule (Brodley & Utgoff, 1995). A recent modification to this procedure (Brodley, 1995) addresses the problem that the weights found by this rule depend on the order in which the instances are presented; a poor ordering can lead to an inaccurate classifier. To minimize this problem, the thermal training procedure is applied ten times, using a different ordering for the instances each time. This produces ten LM's, each with a different set of weights. The LM that maximizes the information-gain ratio metric is then chosen.

## 4.4 Effect of Filtering on Classification Accuracy

In Table 3 we show the accuracy for the land cover data of the classifiers formed by each of the three algorithms tested using no filter (None), a single-algorithm filter (SF)[7], a majority vote filter (MF), and a consensus filter (CF). The first row reports the noise rate used to corrupt the data. Note that for this dataset the percentage of the entire training set that is corrupted for a noise rate of $x$ will be less than $x\%$ because only some pairs of classes are considered problematic. The actual percentage of corrupted training data is reported in the second row of the table.

When no noise is introduced, filtering did not make a significant difference for any of the methods on this dataset. Since the original data is not guaranteed to be noise free, we have no way to evaluate whether filtering improves the *true* classification accuracy using the test data available here.

---

7. In all of the tables, SF refers to the single algorithm filter when the same learning algorithm is used to form both the filter and the final classifier.

For noise levels up to 20%, when given data from a majority filter, all methods were able to retain close to their *base-line accuracy*, which we define to be that obtained for the case of 0% noise and no filtering. For noise levels of 30% and 40%, filtering improves accuracy for all three algorithms, with majority filtering performing slightly better than consensus or single-algorithm filtering. For this dataset, the best classification methods were 1-NN and decision trees, although at a noise level of 40% all three algorithms achieved comparable accuracies. In Table 13 (see Appendix) we show the results of a paired *t*-test comparing not filtering (None) to each of the filtering methods. The table reports the p-value, which is the probability that the difference in the two sample means is due to chance.[8] Figure 5 shows a graph of the accuracy values using a 1-NN as the final classifier of the filtered data. Note that the curve labeled SF refers to the results from using a single algorithm filter constructed via the same learning algorithm as the final classifier.

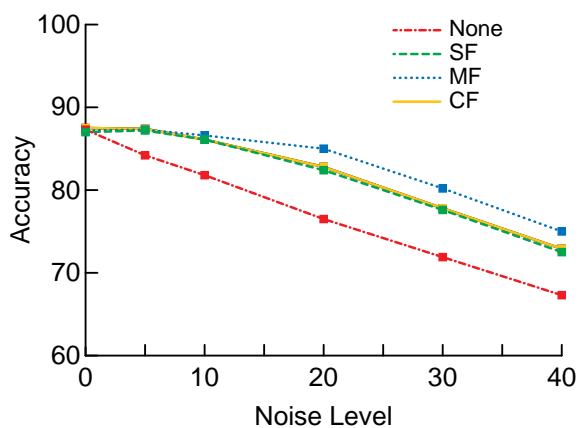

Figure 5: Accuracy of the land cover data for a 1-NN.

For three of the remaining four datasets we show graphs reporting the accuracy of each of the four filtering methods (none, single-algorithm, majority and consensus) in conjunction with a final classifier, selected by choosing for each dataset the most accurate of the three learning algorithms when run without a filter and without injected noise. For the fire dataset, we chose to show the results for a decision tree because these results possess the largest difference in accuracy between filtering and not filtering. The full table of results for each dataset can be found in Tables 13-21 in the Appendix.

For the credit data (Figure 6), the linear machine is a better learning bias than either the 1-NN or the decision tree as evidenced by its higher base-level accuracy (83.5 versus 78.1 and 77.6). In this case applying a single-algorithm or majority filter leads to slightly better results than a consensus filter for noise levels above 5%. At noise levels of 30% and higher, filtering ceases to improve classification accuracy. At 40% noise, it is unlikely that any of the filtering methods could improve accuracy because insufficient high quality training data is available to build an accurate filter. Indeed, at 40% noise, the use of a consensus filter yields lower accuracy relative to not filtering, as the biases of decision trees and 1-NN lead the method to eliminate many good instances (we will expand this point when we discuss the error rates of the filters in Section 4.7.)

---

8. These significance results should be considered optimistic as Dietterich (1998) has illustrated that a paired *t*-test has an elevated Type 1 error. Moreover, by running ten trials, each trial with a random partition into train and test sets, we have violated the assumption that the test sets are independent.

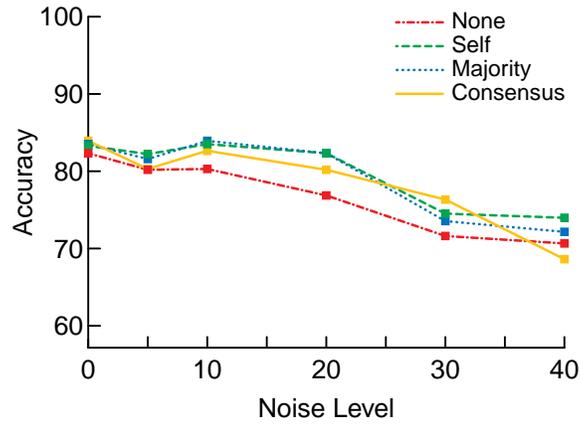

Figure 6: Accuracy of the credit data for a linear machine.

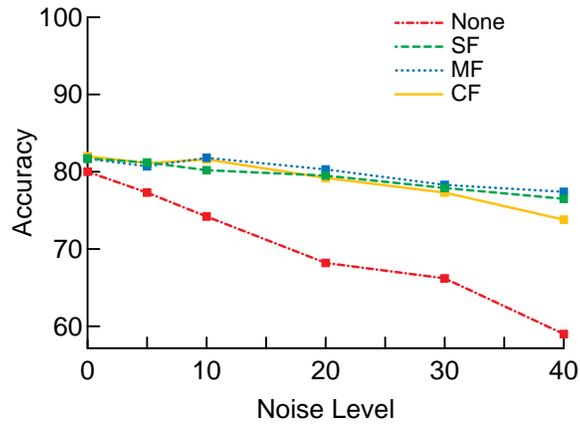

Figure 7: Accuracy of the road segmentation data for a decision tree.

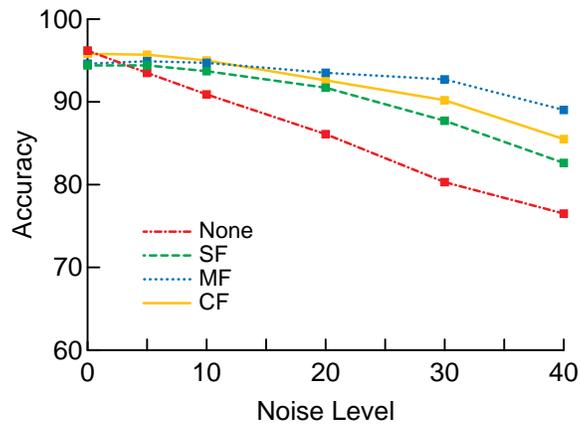

Figure 8: Accuracy of the scene segmentation data for a 1-NN.

Figures 7 and 8 show the results of the road segmentation and scene segmentation respectively. These two datasets are similar in that they were labeled using the same visual labeling process (described in Section 4.1.3), but differ in the features measured and the set of classes. For the road segmentation data, all three filtering methods perform comparably and in each case substantially improve accuracy relative to not filtering. For this dataset, using a decision tree as a final classifier yielded slightly better performance than a 1-NN and we show results for this method. For the scene segmentation data, a 1-NN was a slightly better final classifier as shown in Table 18. For this dataset, at noise levels of 30% and 40% the majority filter performs better than each of the others and retains base-line accuracy at 30% noise. The improvement in accuracy from filtering for each of these two segmentation datasets can be attributed to their class separability. Specifically, it is relatively easy to spot outliers because three of the features in each dataset measure color, and many of the classes are well-separated in spectral space.

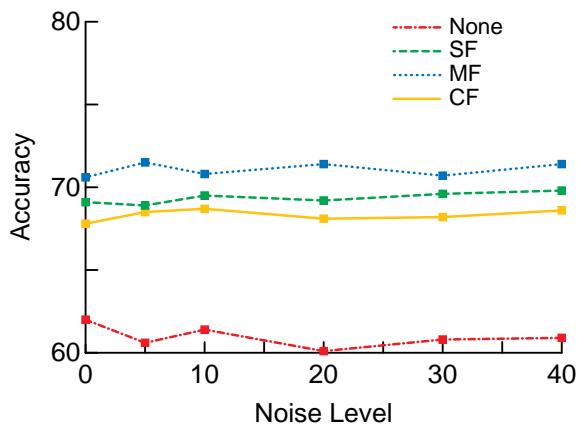

Figure 9: Accuracy of the fire severity data for a decision tree.

The results for the fire severity dataset differ substantially from the other four datasets. Figure 9 shows the results for the decision tree algorithm across the six noise levels. For this dataset, filtering improves classification accuracy for 0% noise. Recall that noise was introduced among classes 2-8, but not in class 1. On closer investigation, we discovered that when applying filtering to the original dataset (0% added noise) almost all of the instances from classes 2-8 were tagged as noisy by all three filter methods. Table 4 shows a comparison of the distribution of instances in the original dataset and of the instances left after a majority filter has been applied. We examined the misclassification matrix for the original data and found that about half of the errors resulted from classifying instances labeled 2-8 as class 1 and the other half were from classifying class $x$ as class $y$, where $x \neq y$ and $2 \leq x, y \leq 8$. After filtering, the dataset contains a preponderance of instances belonging to class 1. Such an uneven distribution of classes results in classifiers biased toward classifying every instance as class 1, which in the original distribution of the instances (maintained in the uncorrupted test instances) is roughly 70%. Introducing more noise into classes 2-8 did not change this behavior and therefore, the accuracy curves (see Figure 9) remain flat across all noise levels. In the traditional use of this dataset (predicting fire versus no fire) accuracies of 70.6% were observed for regression prediction, while for the task of predicting low (1-3) versus high risk days (4-8) the accuracy was 87.2% (Krusel & Dowe, 1993). In summary, for this dataset, we conjecture that either the features are inadequate to discriminate classes 2-8, or the labels in the original data contain a degree of subjectivity that makes it impossible to create an accurate filter.

| | 1 | 2 | 3 | 4 | 5 | 6 | 7 | 8 |
|---|---|---|---|---|---|---|---|---|
| Original Dataset | 2481.0 | 373.0 | 273.0 | 174.0 | 101.0 | 35.0 | 9.0 | 7.0 |
| Majority Filter | 2095.5 | 23.7 | 16.2 | 4.7 | 2.4 | 1.9 | 0.0 | 0.1 |

Table 4: Class distribution for the fire severity dataset with and without filtering averaged over ten runs

| | Land Cover | | Credit | | Road | | Scene | | Fire Severity | |
|---|---|---|---|---|---|---|---|---|---|---|
| Noise | None | CF | None | CF | None | CF | None | CF | None | CF |
| 0 | 187.7 | 121.4 | 76.1 | 48.2 | 177.9 | 93.6 | 45.3 | 40.5 | 575.7 | 177.4 |
| 5 | 271.4 | 129.3 | 89.5 | 55.8 | 234.2 | 93.7 | 98.1 | 42.8 | 576.9 | 178.8 |
| 10 | 321.5 | 144.9 | 99.5 | 58.3 | 281.1 | 98.6 | 138.2 | 51.9 | 581.9 | 173.7 |
| 20 | 401.6 | 195.1 | 114.2 | 70.3 | 352.3 | 111.9 | 187.3 | 76.2 | 586.5 | 162.8 |
| 30 | 447.1 | 234.4 | 123.4 | 79.9 | 389.0 | 121.3 | 230.3 | 98.1 | 594.1 | 156.8 |
| 40 | 467.7 | 267.8 | 126.6 | 83.1 | 432.6 | 132.5 | 255.4 | 116.3 | 589.7 | 153.1 |

Table 5: Tree size – number of leaves

## 4.5 Effect of Filtering on Tree Size

Applying filters to the training data leads to substantially smaller decision trees. Table 5 reports the number of leaves in decision trees produced from the consensus filtered and the unfiltered data.[9] For 0-5% noise, the filtered data creates trees with fewer leaves than trees estimated from the original dataset. For the road segmentation and fire severity datasets, even at 40% noise, the trees produced from the filtered data have fewer leaves than the one produced from the original dataset at 0% noise. This effect was also observed by John (1995) and attributed to *Robust C4.5's* ability to remove "confusing" instances from the training data, thereby reducing the size of the learned decision trees. Oates and Jensen (1997) showed empirically that for many datasets there is a linear relationship between tree size and the number of training instances – randomly increasing the number of training instances has the effect of increasing tree size even when pruning is applied.[10] Their analysis of *Robust C4.5* shows that 41.67% of the decrease in tree size is attributable to reduction

9. The results of paired t-test comparing the number of leaves with and without filtering, shows that the difference in each case is significant as measured at the p=0.01 level.

10. This relationship was found to hold to varying degrees for five different pruning methods.

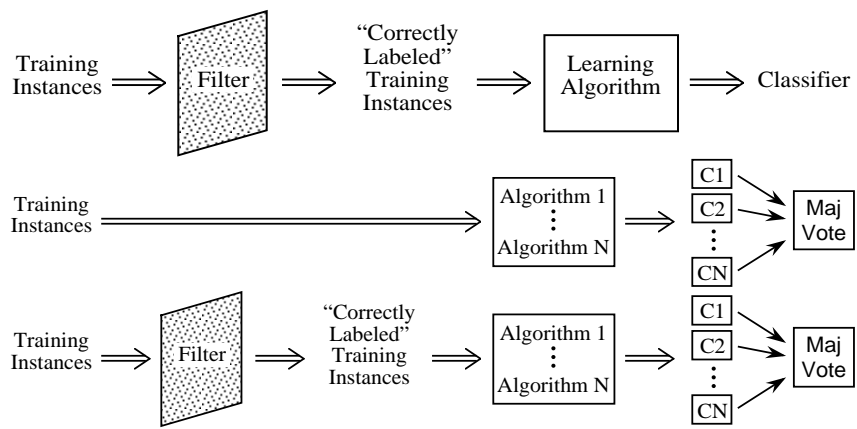

Figure 10: Models compared

the size of the training set. The remainder is due to the removal of uninformative instances (noise).

A second trend in tree size is apparent in the results presented in Table 5: as the noise level increases the size of the trees formed from unfiltered data grows more quickly than the size of the trees formed from filtered data. This reinforces the well-known phenomenon that noise in the class labels increases the size of a decision tree. An exception to this general trend is observed for the fire severity dataset. Note that for this dataset, accuracy and tree size are approximately constant across various noise levels. This results because for each level of noise the method throws out almost the same subset of the instances.

## 4.6 Voting versus Filtering

A hypothesis of interest is whether a majority vote ensemble classifier can be used instead of filtering. To test this hypothesis we formed two majority vote ensemble classifiers: one from the filtered and one from unfiltered data. The majority vote ensemble serves as the *final classifier* and not as the filter (as shown in the bottom two schemes depicted in Figure 10). The resulting classifiers were then used to classify the uncorrupted test data.

The results for the land cover data are shown in Table 6. For each of three methods (None, Majority and Consensus) we compare the accuracy of a majority vote classifier to the 1-NN classifier, which is the most accurate of the three base-level classifiers for this domain. The table includes the p-values of a paired t-test to assess the significance in the difference found between the majority classifier and 1-NN classifier. The majority vote classifier is made up of a 1-NN, a decision tree and a linear machine. We did not use any weighting scheme for combining their votes. For each filtering scheme, the majority vote classifier has equal or better accuracy than the 1-NN classifier with the exception of noise levels of 0-20% for the majority filter. At lower noise levels (0-10%) filtering does not have a large impact on the accuracy of the majority vote classifier. However, at higher noise levels (20-30%), both majority and consensus filtering improve the majority vote classifier's accuracy over that obtained when no filtering method was applied.

| Noise | No Filter | | | Majority Filter | | | Consensus Filter | | |
|-------|------|------|------|------|------|------|------|------|------|
| Level | MAJ | 1-NN | p | MAJ | 1-NN | p | MAJ | 1-NN | p |
| 0 | 87.3 | 87.3 | 0.872 | 86.3 | 87.2 | 0.053 | 87.1 | 87.5 | 0.470 |
| 5 | 86.1 | 84.2 | 0.000 | 85.7 | 87.3 | 0.000 | 87.0 | 87.4 | 0.474 |
| 10 | 85.6 | 81.8 | 0.000 | 85.3 | 86.6 | 0.404 | 87.0 | 86.1 | 0.054 |
| 20 | 80.1 | 76.5 | 0.002 | 84.6 | 85.0 | 0.113 | 85.4 | 82.8 | 0.001 |
| 30 | 75.9 | 71.9 | 0.006 | 82.0 | 80.2 | 0.001 | 82.0 | 77.8 | 0.000 |
| 40 | 71.5 | 67.3 | 0.005 | 75.9 | 75.0 | 0.363 | 75.6 | 72.9 | 0.000 |

Table 6: Comparison of filtering to voting – land cover data

The results for the remaining four datasets are shown in Tables 22-25 in the Appendix. In one case (credit), an individual method (linear machine) was more accurate than, or approximately equivalent to, the majority vote classifier for every filtering method. For the other three datasets, the majority vote classifier was on average better than the single best classifier. Excluding the scene segmentation data, applying a majority or consensus filter and then building a classifier using the single best algorithm outperformed a majority vote classifier without filtering for noise levels of 10% and higher. Except for the 0% noise case for the road and scene segmentation data, and the 5% case for the scene data, filtering improved the accuracy of the majority vote classifier over not filtering. In addition, in many cases (particularly at higher noise levels) filtering applied with the best individual classifier obtained better accuracies than applying the majority vote classifier to unfiltered data. These last two results demonstrate that for these datasets, majority vote classifiers cannot replace filtering.

## 4.7 Filter Precision

To assess the filters' ability to identify mislabeled instances, we examined the intersection between the set of instances that were corrupted and the set of instances that were tagged as mislabeled. In Figure 2 this is the area $M \cap D$. The results of this analysis for the land-cover data are shown in Table 7. Each row in the table reports the average over the ten runs of the number of instances discarded by each filter $D_{SF}$, $D_{MF}$, $D_{CF}$, the number of instances corrupted in the data $M$, and for each filter the number of instances in the intersection of the set of corrupted data and the set of discarded data. Ideally the set of instances discarded should completely intersect the set of noisy instances. Since we may have noisy instances over and above the number of artificially corrupted instances we cannot know the exact number. Therefore in this analysis we approximate our calculations of precision by assuming that the only noisy instances are those that we explicitly corrupted. In this case we would like the intersection between the instances discarded and the instances corrupted to be 100%. In practice we see that this is not the case. Results for the remaining four datasets are given in Tables 26-29 of the Appendix.

In Tables 8-12 we report estimates of the probabilities that each filter makes $E1$ and $E2$ errors. $P(E1)$ represents the probability of throwing out good data and can be estimated as:

| Noise | Instances Discarded | | | Instances | Instances in Intersection | | |
|-------|----------|----------|----------|-----------|-----------------|-----------------|-----------------|
| Level | $D_{SF}$ | $D_{MF}$ | $D_{CF}$ | Corrupted (M) | M ∩ $D_{SF}$ | M ∩ $D_{MF}$ | M ∩ $D_{CF}$ |
| 5 | 622.7 | 569.0 | 265.5 | 106.3 | 98.4 | 100.6 | 91.9 |
| 10 | 747.1 | 676.3 | 332.0 | 207.1 | 180.9 | 187.7 | 155.2 |
| 20 | 968.8 | 895.2 | 446.4 | 435.2 | 330.7 | 356.7 | 253.4 |
| 30 | 1109.8 | 1093.3 | 501.7 | 663.3 | 444.2 | 475.1 | 298.6 |
| 40 | 1203.0 | 1194.6 | 524.6 | 899.9 | 520.0 | 567.7 | 304.3 |

Table 7: The size of the intersection of discarded and mislabeled datasets - land cover data (SF = 1-NN)

| Noise | Self Filter − 1-NN | | Majority Filter | | Consensus Filter | |
|-------|----------|----------|----------|----------|----------|----------|
| Level | $P(E1)$ | $P(E2)$ | $P(E1)$ | $P(E2)$ | $P(E1)$ | $P(E2)$ |
| 5 | 0.18 | 0.07 | 0.16 | 0.05 | 0.06 | 0.14 |
| 10 | 0.20 | 0.13 | 0.17 | 0.09 | 0.06 | 0.25 |
| 20 | 0.24 | 0.24 | 0.20 | 0.18 | 0.07 | 0.42 |
| 30 | 0.28 | 0.33 | 0.26 | 0.28 | 0.08 | 0.55 |
| 40 | 0.31 | 0.42 | 0.29 | 0.37 | 0.10 | 0.66 |

Table 8: Filter precision - land cover data

$$P(E1) = \frac{Discarded - Intersect}{Total - Corrupted} = \frac{D - M \cap D}{Total - M}$$

$P(E2)$ represents the probability of keeping bad data and can be estimated as:

$$P(E2) = \frac{Corrupted - Intersect}{Corrupted} = \frac{M - M \cap D}{M}$$

For the land-cover data, there are 3063 (90% of 3398) total training instances. Therefore, for a noise level of 5% and the consensus filter,

$$P(E1) = \frac{265.5 - 91.9}{3063 - 106.3} = .06$$

and

$$P(E2) = \frac{106.3 - 91.9}{106.3} = .14$$

Tables 8-12 show similar trends. For these datasets, the results for the consensus filter show that the probability of throwing out good data remains small even for higher noise levels, illustrating that the consensus filter is conservative in discarding data. On the other

| Noise | Self Filter − LM | | Majority Filter | | Consensus Filter | |
|-------|--------|--------|--------|--------|--------|--------|
| Level | $P(E1)$ | $P(E2)$ | $P(E1)$ | $P(E2)$ | $P(E1)$ | $P(E2)$ |
| 5  | 0.17 | 0.23 | 0.20 | 0.22 | 0.07 | 0.47 |
| 10 | 0.18 | 0.23 | 0.22 | 0.25 | 0.07 | 0.50 |
| 20 | 0.20 | 0.30 | 0.25 | 0.32 | 0.07 | 0.59 |
| 30 | 0.26 | 0.40 | 0.31 | 0.46 | 0.09 | 0.72 |
| 40 | 0.33 | 0.48 | 0.40 | 0.51 | 0.12 | 0.76 |

Table 9: Filter precision - credit data

| Noise | Self Filter − Dtree | | Majority Filter | | Consensus Filter | |
|-------|--------|--------|--------|--------|--------|--------|
| Level | $P(E1)$ | $P(E2)$ | $P(E1)$ | $P(E2)$ | $P(E1)$ | $P(E2)$ |
| 5  | 0.24 | 0.16 | 0.23 | 0.14 | 0.11 | 0.24 |
| 10 | 0.27 | 0.21 | 0.24 | 0.20 | 0.11 | 0.30 |
| 20 | 0.35 | 0.29 | 0.30 | 0.28 | 0.13 | 0.41 |
| 30 | 0.42 | 0.39 | 0.37 | 0.37 | 0.16 | 0.52 |
| 40 | 0.54 | 0.46 | 0.49 | 0.44 | 0.20 | 0.59 |

Table 10: Filter precision - road segmentation data

| Noise | Self Filter − 1-NN | | Majority Filter | | Consensus Filter | |
|-------|--------|--------|--------|--------|--------|--------|
| Level | $P(E1)$ | $P(E2)$ | $P(E1)$ | $P(E2)$ | $P(E1)$ | $P(E2)$ |
| 5  | 0.06 | 0.07 | 0.05 | 0.05 | 0.01 | 0.10 |
| 10 | 0.09 | 0.15 | 0.06 | 0.09 | 0.01 | 0.21 |
| 20 | 0.14 | 0.25 | 0.08 | 0.18 | 0.01 | 0.33 |
| 30 | 0.18 | 0.33 | 0.11 | 0.25 | 0.02 | 0.44 |
| 40 | 0.22 | 0.42 | 0.15 | 0.34 | 0.02 | 0.55 |

Table 11: Filter precision - scene segmentation data

| Noise | Self Filter − Dtree | | Majority Filter | | Consensus Filter | |
|-------|--------|--------|--------|--------|--------|--------|
| Level | $P(E1)$ | $P(E2)$ | $P(E1)$ | $P(E2)$ | $P(E1)$ | $P(E2)$ |
| 5  | 0.36 | 0.14 | 0.37 | 0.08 | 0.22 | 0.27 |
| 10 | 0.36 | 0.14 | 0.37 | 0.12 | 0.22 | 0.26 |
| 20 | 0.35 | 0.25 | 0.36 | 0.21 | 0.21 | 0.36 |
| 30 | 0.33 | 0.34 | 0.34 | 0.30 | 0.20 | 0.41 |
| 40 | 0.33 | 0.42 | 0.33 | 0.40 | 0.19 | 0.50 |

Table 12: Filter precision - fire severity data

hand, the results illustrate that the probability of a consensus filter electing to keep bad data is larger than the majority vote filter's for each noise level across all data sets. Indeed for a noise level of 40%, the CF has a 66% (land cover), 76% (credit), 59% (road), 55% (scene), and 50% (fire) chance of retaining mislabeled instances. For the majority filter, the chance of making $E1$ and $E2$ errors is more equal. Excluding the credit data at 40% noise, the probability of these errors never reaches above 50%. Considering that majority vote performs better than consensus filters for higher noise rates, this shows that a consensus filter's propensity toward making $E2$ errors (retaining bad data) hinders performance more than majority filter's lesser ability to retain good data (i.e., majority makes more $E1$ errors).

The fire dataset has a very different profile. The probability of throwing out good data remains almost constant across the different noise levels. This is because for this dataset "good" instances appear to have a high level of noise. On the other hand, the probability that a filter will retain bad data rises as the noise level increases.

If one has a lot of data, then an elevated $E1$ error is probably less of a hindrance than an elevated $E2$ error; i.e., throwing out good data, when you have a lot is less costly than retaining bad data. Of course, one would like to insure that one is not throwing out exceptions.

## 5. Conclusions and Future Directions

This article presents a procedure for identifying mislabeled instances. The results of an empirical evaluation demonstrated that filtering improves classification accuracy for datasets that possess labeling errors. Filtering allowed accuracies near to the base-line accuracy to be retained for noise levels up to 20% for all datasets, and up to 30% for two datasets (the road and scene segmentation datasets). Our experiments show that as the noise level increases, the ability of the method to retain the baseline accuracy decreases. Moreover, as illustrated by the fire severity dataset, if the method starts with data that is overly noisy, it cannot form an accurate filter. A comparison of voting to filtering illustrated that the majority vote classifier performed better than the individual classifiers, but that it cannot replace filtering when data are noisy. Our results show that the best approach is to combine filtering and voting. An evaluation of the precision of filtering illustrated that consensus filters are conservative in throwing away good data at the expense of keeping mislabeled data, whereas for majority vote filters the probability of throwing out good data and the probability of retaining bad data are more even. Because majority vote filters perform better on average than consensus filters this shows that retaining bad data hinders performance more than throwing out good data for these datasets. This trend is particularly important when one has an abundance of data.

The issue of determining whether or not to apply filtering to a given data set must be considered. For the work described here, the data were artificially corrupted. Therefore the nature and magnitude of the labeling errors were known a priori. Unfortunately, this type of information is rarely known for most "real world" applications. In some situations, it may be possible to use domain knowledge to estimate the amount of label noise in a dataset. For situations where this knowledge is not available, the conservative nature of the consensus filter dictates that relatively few instances will be discarded for data sets with low levels of labeling error. Therefore, the application of this method to relatively noise free datasets should not significantly impact the performance of the final classification procedure.

A future direction of this research will be to extend the filter approach to *correct* labeling errors in training data. For example, one way to do this might be to relabel instances if the consensus class is different than the observed class. Instances for which the consensus filter predicts two or more classes would still be discarded. This direction is particularly important because of the paucity of high quality training data available for many applications.

A danger in automatically removing instances that cannot be correctly classified is that they might be exceptions to the general rule. When an instance is an exception to the general case, it can appear as though it is incorrectly labeled. When applying techniques to identify and eliminate noisy instances, one wants to avoid discarding correctly labeled exceptions. Therefore a key question in improving data quality is how to distinguish exceptions from noise. One solution to this problem might be to create diagnostics that look at the *way* in which an instance is misclassified in order to determine if it is an exception or an error. We plan to investigate whether with limited feedback, one can learn to distinguish exceptions from noise based on their classification behavior and input feature values.

The experiments described in this paper have been confined to introducing noise into the data in a manner that is natural for the particular domain. This was necessary, because for the datasets used we had no way of ensuring a noise-free validation test set. A key focus of future work will be to generate noise free validation data to test our method on the original data set. We are currently working on obtaining noise free validation data for the land cover classification task.

## Acknowledgments

We would like to thank Ruth DeFries for supplying the NDVI dataset, and Bruce Draper for providing the class confusions for the scene and road segmentation data. We thank Katie Wilson for her comments. We thank the anonymous reviewers and our editor for their careful reading and excellent suggestions. Carla Brodley's research was supported by NSF IIS-9733573 and NASA under grant # NAG5-6971. Mark Friedl's research was supported by NASA under grant # NAG5-7218.

## Appendix A. Additional Results

Table 13 reports the results of a paired *t*-test for the land-cover classification dataset. Tables 14 to 21 report the classification accuracy, the sample standard deviation and the results of a paired t-test for the credit risk, road and scene segmentation, and fire severity datasets. Tables 22 to 25 show the results of a comparison of majority vote classification to filtering, for the credit risk, road and scene segmentation, and fire severity datasets. Tables 26 to 29 show the precision of the filtering methods for the credit risk, road and scene segmentation, and fire severity datasets.

| Noise Level | | 0 | 5 | 10 | 20 | 30 | 40 |
|---|---|---|---|---|---|---|---|
| 1-NN | SF | 0.446 | 0.000 | 0.000 | 0.000 | 0.000 | 0.000 |
| | MF | 0.752 | 0.000 | 0.000 | 0.000 | 0.000 | 0.000 |
| | CF | 0.406 | 0.000 | 0.000 | 0.000 | 0.000 | 0.000 |
| LM | SF | 0.561 | 0.104 | 0.586 | 0.003 | 0.153 | 0.011 |
| | MF | 0.268 | 0.019 | 0.081 | 0.000 | 0.034 | 0.000 |
| | CF | 0.046 | 0.023 | 0.107 | 0.001 | 0.069 | 0.017 |
| D-Tree | SF | 0.070 | 0.280 | 0.000 | 0.000 | 0.000 | 0.000 |
| | MF | 0.158 | 0.757 | 0.001 | 0.000 | 0.000 | 0.000 |
| | CF | 0.934 | 0.008 | 0.001 | 0.000 | 0.000 | 0.000 |

Table 13: Comparison of filtering to not filtering (statistical significance) – land cover data

| Noise Level | | 0 | 5 | 10 | 20 | 30 | 40 |
|---|---|---|---|---|---|---|---|
| Actual Noise | | 0 | 5.5 | 10.0 | 19.6 | 32.4 | 43.9 |
| 1-NN | None | $78.1 \pm 5.2$ | $75.1 \pm 6.1$ | $71.9 \pm 4.9$ | $65.4 \pm 6.9$ | $62.9 \pm 6.8$ | $59.9 \pm 5.4$ |
| | SF | $77.9 \pm 4.7$ | $75.4 \pm 5.3$ | $75.6 \pm 4.6$ | $71.9 \pm 9.2$ | $67.9 \pm 5.5$ | $60.6 \pm 3.7$ |
| | MF | $81.5 \pm 5.3$ | $78.4 \pm 6.5$ | $78.4 \pm 4.2$ | $75.3 \pm 6.7$ | $71.0 \pm 6.1$ | $63.7 \pm 4.0$ |
| | CF | $81.5 \pm 4.5$ | $77.6 \pm 5.5$ | $75.6 \pm 5.1$ | $73.2 \pm 8.0$ | $69.0 \pm 7.6$ | $63.7 \pm 5.4$ |
| LM | None | $83.5 \pm 4.8$ | $81.5 \pm 3.3$ | $81.6 \pm 4.3$ | $78.4 \pm 4.2$ | $73.5 \pm 5.7$ | $72.6 \pm 5.2$ |
| | SF | $84.4 \pm 5.5$ | $83.4 \pm 3.6$ | $84.6 \pm 5.1$ | $83.5 \pm 3.0$ | $76.2 \pm 5.4$ | $75.7 \pm 4.0$ |
| | MF | $84.6 \pm 3.7$ | $82.8 \pm 3.5$ | $85.0 \pm 2.3$ | $83.5 \pm 4.7$ | $75.3 \pm 7.2$ | $74.0 \pm 5.8$ |
| | CF | $85.0 \pm 3.7$ | $81.6 \pm 4.8$ | $83.8 \pm 4.7$ | $81.5 \pm 4.6$ | $77.9 \pm 6.8$ | $70.7 \pm 7.4$ |
| D-Tree | None | $77.6 \pm 6.0$ | $75.9 \pm 6.3$ | $70.7 \pm 6.0$ | $68.5 \pm 6.9$ | $64.1 \pm 6.4$ | $58.5 \pm 5.2$ |
| | SF | $75.6 \pm 5.5$ | $73.7 \pm 3.9$ | $76.0 \pm 4.9$ | $72.4 \pm 4.0$ | $66.0 \pm 6.3$ | $55.7 \pm 7.9$ |
| | MF | $74.9 \pm 3.9$ | $79.0 \pm 5.4$ | $78.2 \pm 4.9$ | $75.7 \pm 4.2$ | $69.9 \pm 6.9$ | $64.6 \pm 5.3$ |
| | CF | $81.3 \pm 4.2$ | $78.1 \pm 4.8$ | $78.7 \pm 5.5$ | $74.1 \pm 6.5$ | $65.6 \pm 7.8$ | $61.9 \pm 6.4$ |

Table 14: Classification accuracy – credit data

| Noise Level | | 0 | 5 | 10 | 20 | 30 | 40 |
|---|---|---|---|---|---|---|---|
| 1-NN | SF | 0.907 | 0.820 | 0.023 | 0.006 | 0.037 | 0.660 |
| | MF | 0.020 | 0.022 | 0.000 | 0.000 | 0.005 | 0.103 |
| | CF | 0.000 | 0.022 | 0.002 | 0.000 | 0.001 | 0.000 |
| LM | SF | 0.460 | 0.158 | 0.138 | 0.002 | 0.153 | 0.044 |
| | MF | 0.550 | 0.171 | 0.012 | 0.001 | 0.510 | 0.499 |
| | CF | 0.186 | 0.909 | 0.267 | 0.074 | 0.068 | 0.477 |
| D-Tree | SF | 0.301 | 0.290 | 0.023 | 0.131 | 0.207 | 0.320 |
| | MF | 0.103 | 0.128 | 0.000 | 0.013 | 0.065 | 0.009 |
| | CF | 0.016 | 0.322 | 0.001 | 0.008 | 0.405 | 0.125 |

Table 15: Comparison of filtering to not filtering (statistical significance) – credit data

| Noise Level | | 0 | 5 | 10 | 20 | 30 | 40 |
|---|---|---|---|---|---|---|---|
| Actual Noise | | 0.0 | 5.1 | 10.6 | 22.6 | 36.5 | 52.5 |
| 1-NN | None | 79.6 ± 1.7 | 76.7 ± 1.7 | 71.7 ± 2.8 | 66.6 ± 2.7 | 62.5 ± 2.3 | 57.2 ± 3.0 |
| | SF | 82.0 ± 1.5 | 80.5 ± 1.8 | 80.9 ± 1.2 | 79.4 ± 1.7 | 77.2 ± 1.6 | 74.5 ± 2.2 |
| | MF | 82.6 ± 1.5 | 81.6 ± 1.5 | 82.2 ± 1.9 | 80.9 ± 1.0 | 79.7 ± 1.8 | 77.2 ± 2.9 |
| | CF | 81.9 ± 1.4 | 81.6 ± 1.3 | 81.2 ± 2.3 | 78.9 ± 1.6 | 76.8 ± 1.1 | 74.5 ± 2.3 |
| LM | None | 76.6 ± 4.5 | 69.7 ± 11.7 | 68.0 ± 6.7 | 61.6 ± 11.1 | 65.9 ± 7.4 | 59.4 ± 5.0 |
| | SF | 76.9 ± 3.5 | 76.2 ± 6.5 | 78.1 ± 2.2 | 73.7 ± 4.5 | 73.9 ± 5.9 | 74.9 ± 3.7 |
| | MF | 77.7 ± 2.0 | 77.2 ± 5.4 | 76.5 ± 3.8 | 78.6 ± 1.9 | 77.2 ± 2.6 | 76.4 ± 2.3 |
| | CF | 78.3 ± 2.6 | 76.1 ± 3.0 | 75.3 ± 4.7 | 74.5 ± 6.1 | 72.9 ± 6.7 | 71.6 ± 4.7 |
| D-Tree | None | 80.0 ± 1.4 | 77.3 ± 2.2 | 74.2 ± 2.5 | 68.2 ± 2.3 | 66.2 ± 3.0 | 59.0 ± 4.3 |
| | SF | 81.7 ± 1.8 | 81.2 ± 1.4 | 80.2 ± 2.3 | 79.5 ± 2.0 | 77.9 ± 2.1 | 76.5 ± 1.8 |
| | MF | 81.7 ± 1.3 | 80.7 ± 1.8 | 81.8 ± 1.4 | 80.3 ± 1.4 | 78.3 ± 1.5 | 77.4 ± 2.0 |
| | CF | 82.0 ± 1.2 | 81.1 ± 1.7 | 81.6 ± 2.7 | 79.2 ± 1.9 | 77.3 ± 1.8 | 73.8 ± 2.0 |

Table 16: Classification accuracy – road segmentation data

| Noise Level | | 0 | 5 | 10 | 20 | 30 | 40 |
|---|---|---|---|---|---|---|---|
| 1-NN | SF | 0.000 | 0.000 | 0.000 | 0.000 | 0.000 | 0.000 |
| | MF | 0.001 | 0.000 | 0.000 | 0.000 | 0.000 | 0.000 |
| | CF | 0.001 | 0.000 | 0.000 | 0.000 | 0.000 | 0.000 |
| LM | SF | 0.856 | 0.197 | 0.001 | 0.002 | 0.054 | 0.000 |
| | MF | 0.548 | 0.105 | 0.009 | 0.001 | 0.001 | 0.000 |
| | CF | 0.212 | 0.123 | 0.010 | 0.004 | 0.042 | 0.000 |
| D-Tree | SF | 0.024 | 0.000 | 0.000 | 0.000 | 0.000 | 0.000 |
| | MF | 0.020 | 0.001 | 0.000 | 0.000 | 0.000 | 0.000 |
| | CF | 0.002 | 0.002 | 0.000 | 0.000 | 0.000 | 0.000 |

Table 17: Comparison of filtering to not filtering (statistical significance) – road segmentation data

| Noise Level | | 0 | 5 | 10 | 20 | 30 | 40 |
| Actual Noise | | 0.0 | 2.8 | 6.0 | 12.5 | 19.5 | 28.0 |
|---|---|---|---|---|---|---|---|
| 1-NN | None | $96.2 \pm 1.1$ | $93.5 \pm 1.4$ | $90.9 \pm 1.8$ | $86.1 \pm 2.4$ | $80.3 \pm 2.0$ | $76.5 \pm 3.5$ |
| | SF | $94.4 \pm 1.2$ | $94.4 \pm 1.4$ | $93.7 \pm 1.6$ | $91.7 \pm 1.5$ | $87.7 \pm 1.8$ | $82.6 \pm 3.3$ |
| | MF | $94.6 \pm 1.4$ | $94.9 \pm 1.1$ | $94.7 \pm 1.2$ | $93.5 \pm 1.8$ | $92.7 \pm 3.0$ | $89.0 \pm 3.1$ |
| | CF | $95.8 \pm 1.1$ | $95.7 \pm 1.1$ | $95.0 \pm 1.6$ | $92.6 \pm 1.6$ | $90.2 \pm 2.7$ | $85.5 \pm 2.5$ |
| LM | None | $90.2 \pm 2.4$ | $91.0 \pm 1.3$ | $90.1 \pm 2.0$ | $90.0 \pm 3.2$ | $88.2 \pm 2.2$ | $86.2 \pm 2.3$ |
| | SF | $89.8 \pm 1.6$ | $89.2 \pm 2.2$ | $89.8 \pm 1.9$ | $90.6 \pm 2.0$ | $90.3 \pm 1.8$ | $89.0 \pm 1.7$ |
| | MF | $89.8 \pm 2.1$ | $90.3 \pm 1.4$ | $90.3 \pm 1.7$ | $90.6 \pm 1.9$ | $89.8 \pm 1.8$ | $90.7 \pm 2.2$ |
| | CF | $91.0 \pm 1.8$ | $90.2 \pm 1.5$ | $91.0 \pm 2.1$ | $91.0 \pm 2.3$ | $91.0 \pm 2.5$ | $89.5 \pm 2.6$ |
| D-Tree | None | $95.8 \pm 1.6$ | $94.5 \pm 1.4$ | $91.4 \pm 1.6$ | $86.0 \pm 3.2$ | $81.9 \pm 1.6$ | $77.6 \pm 3.4$ |
| | SF | $94.2 \pm 2.4$ | $93.9 \pm 1.6$ | $94.1 \pm 2.2$ | $92.2 \pm 2.3$ | $89.0 \pm 3.1$ | $85.4 \pm 4.5$ |
| | MF | $94.8 \pm 1.6$ | $94.4 \pm 1.9$ | $94.5 \pm 1.7$ | $93.2 \pm 2.5$ | $92.6 \pm 2.2$ | $89.5 \pm 2.8$ |
| | CF | $95.5 \pm 1.1$ | $94.9 \pm 1.3$ | $94.7 \pm 1.5$ | $92.4 \pm 2.2$ | $89.5 \pm 3.1$ | $86.9 \pm 2.1$ |

Table 18: Classification accuracy – scene segmentation data

| Noise Level | | 0 | 5 | 10 | 20 | 30 | 40 |
|---|---|---|---|---|---|---|---|
| 1-NN | SF | 0.002 | 0.141 | 0.000 | 0.000 | 0.000 | 0.000 |
| | MF | 0.004 | 0.007 | 0.000 | 0.000 | 0.000 | 0.000 |
| | CF | 0.042 | 0.000 | 0.000 | 0.000 | 0.000 | 0.000 |
| LM | SF | 0.532 | 0.043 | 0.496 | 0.594 | 0.014 | 0.004 |
| | MF | 0.506 | 0.200 | 0.771 | 0.423 | 0.049 | 0.001 |
| | CF | 0.105 | 0.042 | 0.221 | 0.346 | 0.011 | 0.022 |
| D-Tree | SF | 0.020 | 0.169 | 0.005 | 0.001 | 0.000 | 0.000 |
| | MF | 0.046 | 0.818 | 0.001 | 0.000 | 0.000 | 0.000 |
| | CF | 0.343 | 0.193 | 0.000 | 0.000 | 0.000 | 0.000 |

Table 19: Comparison of filtering to not filtering (statistical significance) – scene segmentation data

| Noise Level | | 0 | 5 | 10 | 20 | 30 | 40 |
|---|---|---|---|---|---|---|---|
| Actual Noise | | | 1.4 | 3.1 | 6.9 | 11.6 | 17.9 |
| 1-NN | None | $60.4 \pm 1.7$ | $60.4 \pm 1.7$ | $60.2 \pm 1.6$ | $60.2 \pm 2.0$ | $59.8 \pm 1.8$ | $59.9 \pm 2.2$ |
| | SF | $68.1 \pm 0.8$ | $68.2 \pm 1.5$ | $68.2 \pm 1.6$ | $68.3 \pm 1.3$ | $68.5 \pm 0.9$ | $69.1 \pm 1.0$ |
| | MF | $69.9 \pm 0.9$ | $70.4 \pm 0.8$ | $70.3 \pm 0.8$ | $70.8 \pm 1.0$ | $70.9 \pm 0.8$ | $70.9 \pm 1.0$ |
| | CF | $66.6 \pm 1.1$ | $67.1 \pm 1.4$ | $67.5 \pm 1.1$ | $67.4 \pm 1.7$ | $67.7 \pm 1.2$ | $67.9 \pm 1.0$ |
| LM | None | $64.7 \pm 2.8$ | $63.3 \pm 2.4$ | $63.0 \pm 2.9$ | $62.2 \pm 3.0$ | $66.2 \pm 1.6$ | $65.0 \pm 1.9$ |
| | SF | $63.3 \pm 4.2$ | $65.0 \pm 4.4$ | $63.7 \pm 2.0$ | $62.6 \pm 5.2$ | $63.7 \pm 4.2$ | $64.1 \pm 3.3$ |
| | MF | $67.8 \pm 3.1$ | $65.7 \pm 7.2$ | $66.9 \pm 5.6$ | $64.4 \pm 8.7$ | $67.8 \pm 3.6$ | $67.9 \pm 5.2$ |
| | CF | $66.1 \pm 3.7$ | $66.3 \pm 2.6$ | $66.8 \pm 1.8$ | $67.1 \pm 2.8$ | $67.7 \pm 2.4$ | $67.1 \pm 2.9$ |
| D-Tree | None | $62.0 \pm 1.6$ | $60.6 \pm 1.7$ | $61.4 \pm 1.8$ | $60.1 \pm 1.2$ | $60.8 \pm 1.2$ | $60.9 \pm 2.4$ |
| | SF | $69.1 \pm 1.3$ | $68.9 \pm 1.3$ | $69.5 \pm 1.1$ | $69.2 \pm 1.0$ | $69.6 \pm 1.2$ | $69.8 \pm 1.0$ |
| | MF | $70.6 \pm 0.9$ | $71.5 \pm 1.1$ | $70.8 \pm 1.1$ | $71.4 \pm 0.9$ | $70.7 \pm 1.5$ | $71.4 \pm 0.6$ |
| | CF | $67.8 \pm 1.3$ | $68.5 \pm 1.8$ | $68.7 \pm 1.6$ | $68.1 \pm 1.6$ | $68.2 \pm 1.3$ | $68.6 \pm 1.0$ |

Table 20: Classification accuracy – fire data

| Noise Level | | 0 | 5 | 10 | 20 | 30 | 40 |
|---|---|---|---|---|---|---|---|
| 1-NN | SF | 0.000 | 0.000 | 0.000 | 0.000 | 0.000 | 0.000 |
| | MF | 0.000 | 0.000 | 0.000 | 0.000 | 0.000 | 0.000 |
| | CF | 0.000 | 0.000 | 0.000 | 0.000 | 0.000 | 0.000 |
| LM | SF | 0.245 | 0.214 | 0.576 | 0.770 | 0.096 | 0.494 |
| | MF | 0.021 | 0.329 | 0.127 | 0.433 | 0.218 | 0.188 |
| | CF | 0.232 | 0.052 | 0.002 | 0.000 | 0.146 | 0.168 |
| D-Tree | SF | 0.000 | 0.000 | 0.000 | 0.000 | 0.000 | 0.000 |
| | MF | 0.000 | 0.000 | 0.000 | 0.000 | 0.000 | 0.000 |
| | CF | 0.000 | 0.000 | 0.000 | 0.000 | 0.000 | 0.000 |

Table 21: Comparison of filtering to not filtering (statistical significance) − fire data

| Noise | No filter | | | Majority filter | | | Consensus filter | | |
|---|---|---|---|---|---|---|---|---|---|
| Level | MAJ | LM | p | MAJ | LM | p | MAJ | LM | p |
| 0 | 82.8 | 83.5 | 0.380 | 83.7 | 84.6 | 0.247 | 84.1 | 85.0 | 0.181 |
| 5 | 82.2 | 81.5 | 0.563 | 81.8 | 82.8 | 0.385 | 82.1 | 81.6 | 0.422 |
| 10 | 79.0 | 81.6 | 0.073 | 82.9 | 85.0 | 0.113 | 82.5 | 83.8 | 0.115 |
| 20 | 75.7 | 78.4 | 0.117 | 80.9 | 83.5 | 0.027 | 80.3 | 81.5 | 0.383 |
| 30 | 70.7 | 73.5 | 0.183 | 75.9 | 75.3 | 0.399 | 73.5 | 77.9 | 0.036 |
| 40 | 68.2 | 72.6 | 0.056 | 70.0 | 74.0 | 0.022 | 67.2 | 70.7 | 0.020 |

Table 22: Comparison of filtering to voting − credit data

| Noise | No filter | | | Majority filter | | | Consensus filter | | |
|---|---|---|---|---|---|---|---|---|---|
| Level | MAJ | D-Tree | p | MAJ | D-Tree | p | MAJ | D-Tree | p |
| 0 | 82.8 | 80.0 | 0.001 | 82.1 | 81.7 | 0.331 | 83.2 | 82.0 | 0.012 |
| 5 | 81.7 | 77.3 | 0.000 | 82.2 | 80.7 | 0.006 | 83.3 | 81.1 | 0.001 |
| 10 | 79.0 | 74.2 | 0.000 | 82.2 | 81.8 | 0.235 | 82.8 | 81.6 | 0.085 |
| 20 | 74.8 | 68.2 | 0.000 | 81.5 | 80.3 | 0.013 | 81.5 | 79.2 | 0.004 |
| 30 | 72.6 | 66.2 | 0.000 | 80.5 | 78.3 | 0.001 | 79.2 | 77.3 | 0.022 |
| 40 | 67.7 | 59.0 | 0.000 | 79.5 | 77.4 | 0.001 | 77.8 | 73.8 | 0.000 |

Table 23: Comparison of filtering to voting − road segmentation data

| Noise | No filter | | | Majority filter | | | Consensus filter | | |
|-------|------|------|-------|------|------|-------|------|------|-------|
| Level | MAJ | 1-NN | p | MAJ | 1-NN | p | MAJ | 1-NN | p |
| 0 | 96.8 | 96.2 | 0.078 | 95.4 | 94.6 | 0.136 | 96.5 | 95.8 | 0.042 |
| 5 | 96.8 | 93.5 | 0.000 | 95.5 | 94.9 | 0.115 | 96.3 | 95.7 | 0.123 |
| 10 | 95.2 | 90.9 | 0.000 | 95.7 | 94.7 | 0.032 | 96.5 | 95.0 | 0.003 |
| 20 | 93.1 | 86.1 | 0.000 | 94.9 | 93.5 | 0.012 | 95.5 | 92.6 | 0.000 |
| 30 | 89.9 | 80.3 | 0.000 | 94.8 | 92.7 | 0.007 | 94.5 | 90.2 | 0.000 |
| 40 | 86.8 | 76.5 | 0.000 | 93.1 | 89.0 | 0.000 | 91.8 | 85.5 | 0.000 |

Table 24: Comparison of filtering to voting – scene segmentation data

| Noise | No filter | | | Majority filter | | | Consensus filter | | |
|-------|------|--------|-------|------|--------|-------|------|--------|-------|
| Level | MAJ | D-Tree | p | MAJ | D-Tree | p | MAJ | D-Tree | p |
| 0 | 66.3 | 62.0 | 0.000 | 71.3 | 70.6 | 0.437 | 69.1 | 67.8 | 0.004 |
| 5 | 64.9 | 60.6 | 0.000 | 71.7 | 71.5 | 0.535 | 69.9 | 68.5 | 0.001 |
| 10 | 65.6 | 61.4 | 0.000 | 71.4 | 70.8 | 0.063 | 69.8 | 68.7 | 0.024 |
| 20 | 65.0 | 60.1 | 0.000 | 71.8 | 71.4 | 0.159 | 69.9 | 68.1 | 0.001 |
| 30 | 65.9 | 60.8 | 0.000 | 71.1 | 70.7 | 0.009 | 69.9 | 68.2 | 0.000 |
| 40 | 64.9 | 60.9 | 0.000 | 71.5 | 71.4 | 0.642 | 70.0 | 68.6 | 0.001 |

Table 25: Comparison of filtering to voting – fire data

| Noise | Instances Discarded | | | Instances | Instances in Intersection | | |
|-------|----------|----------|----------|-----------|----------------|----------------|----------------|
| Level | $D_{SF}$ | $D_{MF}$ | $D_{CF}$ | Corrupted (M) | M $\cap$ $D_{SF}$ | M $\cap$ $D_{MF}$ | M $\cap$ $D_{CF}$ |
| 5 | 127.4 | 142.0 | 60.4 | 34.0 | 26.1 | 26.6 | 18.0 |
| 10 | 153.0 | 171.0 | 73.7 | 66.0 | 51.1 | 49.3 | 33.1 |
| 20 | 189.5 | 210.9 | 88.9 | 129.5 | 91.1 | 88.3 | 52.7 |
| 30 | 233.1 | 241.3 | 101.0 | 214.2 | 128.4 | 116.6 | 60.5 |
| 40 | 260.5 | 274.1 | 106.8 | 290.8 | 152.3 | 142.1 | 68.5 |

Table 26: The size of the intersection of discarded and mislabeled datasets - credit data (SF = LM )

| Noise | Instances Discarded | | | Instances | Instances in Intersection | | |
|-------|--------|--------|--------|------------|------------|------------|------------|
| Level | $D_{SF}$ | $D_{MF}$ | $D_{CF}$ | Corrupted (M) | M $\cap$ $D_{SF}$ | M $\cap$ $D_{MF}$ | M $\cap$ $D_{CF}$ |
| 5  | 509.6  | 481.7 | 261.5 | 94.2  | 79.2  | 80.6  | 71.6  |
| 10 | 605.2  | 558.3 | 322.4 | 195.8 | 154.5 | 157.5 | 136.8 |
| 20 | 792.3  | 735.8 | 427.9 | 418.0 | 295.8 | 302.7 | 244.9 |
| 30 | 910.0  | 861.2 | 510.5 | 676.8 | 415.8 | 425.0 | 326.6 |
| 40 | 1002.9 | 976.0 | 575.9 | 974.3 | 530.5 | 544.9 | 400.5 |

Table 27: The size of the intersection of discarded and mislabeled datasets - road segmentation data (SF = D-Tree)

| Noise | Instances Discarded | | | Instances | Instances in Intersection | | |
|-------|--------|--------|--------|------------|------------|------------|------------|
| Level | $D_{SF}$ | $D_{MF}$ | $D_{CF}$ | Corrupted (M) | M $\cap$ $D_{SF}$ | M $\cap$ $D_{MF}$ | M $\cap$ $D_{CF}$ |
| 5  | 183.0 | 152.7 | 75.0  | 57.5  | 53.2  | 54.5  | 51.7  |
| 10 | 286.6 | 228.8 | 122.1 | 124.7 | 105.5 | 113.1 | 99.0  |
| 20 | 455.4 | 364.0 | 198.0 | 259.0 | 194.6 | 213.6 | 172.8 |
| 30 | 571.3 | 490.5 | 254.5 | 405.8 | 270.6 | 305.9 | 228.8 |
| 40 | 663.5 | 604.6 | 297.3 | 582.8 | 339.1 | 386.7 | 262.7 |

Table 28: The size of the intersection of discarded and mislabeled datasets - scene segmentation data (SF = 1-NN)

| Noise | Instances Discarded | | | Instances | Instances in Intersection | | |
|-------|--------|--------|--------|------------|------------|------------|------------|
| Level | $D_{SF}$ | $D_{MF}$ | $D_{CF}$ | Corrupted (M) | M $\cap$ $D_{SF}$ | M $\cap$ $D_{MF}$ | M $\cap$ $D_{CF}$ |
| 5  | 1269.7 | 1301.0 | 797.5 | 47.8  | 41.1  | 43.8  | 34.9  |
| 10 | 1303.5 | 1327.3 | 815.1 | 107.2 | 92.0  | 94.7  | 79.6  |
| 20 | 1300.2 | 1331.1 | 832.3 | 240.0 | 179.4 | 189.0 | 153.0 |
| 30 | 1273.8 | 1317.7 | 844.3 | 400.1 | 266.0 | 280.8 | 234.3 |
| 40 | 1287.1 | 1321.7 | 859.2 | 618.9 | 357.9 | 372.8 | 310.6 |

Table 29: The size of the intersection of discarded and mislabeled datasets - fire data (SF = D-Tree)